\documentclass[9pt,journal]{IEEEtran}

\usepackage{amsmath,amsfonts,bm, bbm}

\def\eqref#1{equation~\ref{#1}}

\def\1{\bm{1}}

\def\rz{{\textnormal{z}}}

\def\rvb{{\mathbf{b}}}

\def\rmU{{\mathbf{U}}}
\def\rmV{{\mathbf{V}}}
\def\rmW{{\mathbf{W}}}

\def\vdelta{{\bm{\delta}}}

\def\vmu{{\bm{\mu}}}

\def\va{{\bm{a}}}
\def\vb{{\bm{b}}}

\def\vg{{\bm{g}}}
\def\vh{{\bm{h}}}

\def\vx{{\bm{x}}}

\def\vz{{\bm{z}}}

\def\mA{{\bm{A}}}

\def\mI{{\bm{I}}}

\def\mK{{\bm{K}}}

\def\mU{{\bm{U}}}

\def\mW{{\bm{W}}}

\def\mSigma{{\bm{\Sigma}}}

\DeclareMathAlphabet{\mathsfit}{\encodingdefault}{\sfdefault}{m}{sl}
\SetMathAlphabet{\mathsfit}{bold}{\encodingdefault}{\sfdefault}{bx}{n}

\def\E{{\mathbb{E}}}

\usepackage{mdframed}

\newtheorem{lm}{Lemma}

\usepackage{accents}
\newlength{\dhatheight}

\usepackage{cite}
\usepackage{lipsum}
\usepackage{adjustbox}    
\usepackage[dvipsnames]{xcolor}
\usepackage[utf8]{inputenc} 
\usepackage[T1]{fontenc}    
\usepackage{hyperref}       
\usepackage{url}            
\usepackage{booktabs}       
\usepackage{amsfonts}       
\usepackage{nicefrac}       
\usepackage[position=top]{subfig}
\usepackage{caption}
\usepackage{tikz}
\usepackage{enumitem}
\usepackage{soul}
\usepackage{tabularx}
\usetikzlibrary{er,positioning}

\definecolor{ao}{rgb}{0.0, 0.5, 0.0}
\definecolor{bblue}{rgb}{0.19, 0.55, 0.91}
\definecolor{bt}{rgb}{1.0, 0.44, 0.37}
\definecolor{blush}{rgb}{0.87, 0.36, 0.51}

\usetikzlibrary{fit,positioning}
\tikzstyle{block} = [rectangle, draw, fill=blue!20, 
    text width=12.8em, text centered, rounded corners, minimum height=4em]
\tikzstyle{line} = [draw, -latex']
\tikzstyle{cloud} = [draw, ellipse,fill=red!20, node distance=3cm,
    minimum height=2em]

\usepackage{algorithm}
\usepackage[noend]{algpseudocode}

\ifCLASSINFOpdf
\else
\fi

\begin{document}

\title{Enhanced Recurrent Neural Tangent Kernels\\ for Non-Time-Series Data}

\author{\IEEEauthorblockN{Anonymous authors}}

\author{
\begin{tabular}[t]{c@{\extracolsep{5em}}c@{\extracolsep{5em}}c@{\extracolsep{5em}}c} 
Sina~Alemohammad  & Randall~Balestriero & Zichao~Wang & Richard~G.~Baraniuk\\
Rice University & Facebook AI Research & Rice University & Rice University\\ 
sa86@rice.edu & rbalestriero@fb.com & zw16@rice.edu & richb@rice.edu
\end{tabular}
}
\maketitle

\begin{abstract}
Kernels derived from deep neural networks (DNNs) in the infinite-width regime provide not only high performance in a range of machine learning tasks but also new theoretical insights into DNN training dynamics and generalization.
In this paper, we extend the family of kernels associated with recurrent neural networks (RNNs), which were previously derived only for simple RNNs, to more complex architectures including bidirectional RNNs and RNNs with average pooling. 
We also develop a fast GPU implementation to exploit the full practical potential of the kernels. 
Though RNNs are typically only applied to time-series data, we demonstrate that classifiers using RNN-based kernels outperform a range of baseline methods on $90$ \emph{non-time-series} datasets from the UCI data repository. 
\end{abstract}


\IEEEpeerreviewmaketitle

\section{Introduction}

Deep neural networks (DNNs) have become the main parametric models used to solve machine learning problems. Among DNNs models, Feed forward networks (FFNs) are proved to be universal approximators \cite{HORNIK1989359,zhou2020universality}, i.e, for any function there exists a FNN that approximates any continuous function to an arbitrary accuracy.

More recent work on {\em infinite-width regime} DNNs, in which the size of the hidden layers extends to infinity, has established a rigorous link between overparametrized DNNs and kernel machines by studying. For example, with random Gaussian parameters, the output of the network is drawn from a Gaussian Process (GP) with an associated Conjugate Kernel (CK) (also known as the NN-GP kernel) in the infinite-width limit \cite{neal1996priors}. Taking the advantage of the equivalence between randomly initialized networks and GPs, \cite{jacot2018neural,lee2019wide} showed that infinite-width DNNs trained with gradient descent converge to a kernel ridge regression predictor with respect to a kernel known as the {\em Neural Tangent Kernel} (NTK).
Each DNN architecture has a distinct CK and NTK which have been derived for different DNN architectures, including convolutions neural networks (CNNs) \cite{arora2019exact,novak2018bayesian}, graph neural networks \cite{du2019graph}, residual networks \cite{huang2020deep,tirer2020kernel}, and attention mechanisms \cite{hron2020infinite}.
Furthermore, \cite{yang2019tensor} and \cite{yang2020tensor} have developed a general methodology to derive the CK and NTK of any DNN architecture with arbitrary building blocks. 

The kernel interpretation of DNNs enables the use of kernel machines operating with DNN-inspired kernels. For example, \cite{Arora2020Harnessing} evaluated the performance of support vector machine (SVM) classifiers with the NTK of multi-layer perceptron (MLP) on 90 datasets from the UCI data repository \cite{yeh1998modeling,czerniak2003application,guyon2004result,cortez2007data,elter2007prediction,little2007exploiting,yeh2007modeling,cortez2009modeling,yeh2009knowledge,bhatt2012planning,tsanas2013objective,santos2015new,set2017semeion,dua2019uci,sikora2019guider,adak2020classification,blachnik2019predicting,glazkova2018automatic,zamora2007behaviour,naranjo2016addressing,alizadehsani2016coronary,liu2017using,dangelo2015artificial,yeh2018building,zieba2013boosted,rozemberczki2021chickenpox,lucas2013failure,karli2007fuzzy,diaz2016reduced,fonollosa2016calibration,martiniano2012application,cassotti2015similarity,groemping2019south,erdem2015baum,abdelhamid2014phishing,banos2014dealing,toth2016miskolcIIS,mallah2013plant,hoseinzade2019cnnpred,rohra2017user,sapsanis2013recognition,cinar2019classification,zamora2014line,johnson2013hybrid}. The NTK of MLP outperformed classical kernels, random forest (RF) and finite-regime trained MLPs. \cite{arora2019exact} reported the best kernel performance on CIFAR-10 using the NTK of a CNN architecture. \cite{li2019enhanced} improved upon the results on CIFAR-10 by using NTK of more complex CNN architectures. Taken together, these studies indicate that DNN-inspired kernels can consistently outperform classical kernels, and in some cases finite trained DNNs. 

In this work, instead of focusing on FNNs, we study Recurrent neural networks (RNNs), which are another powerful class of DNNs that is primarily used for time-series data. RNNs perform sequential operations on an ordered sequence $( \vz^{(1)},\vz^{(2)}, \dots ,\vz^{(T)}  )$ of length $T$ to construct an approximation to the target function. 
Along the lines of research on over-parametrized DNNs, \cite{panigrahi2021learning} recently showed that overparametrized RNNs (with number of parameters orders of magnitude greater than the number of training samples) can approximate all continuous functions on a fixed sequence length of data i.e. they are universal approximators. 
Subsquently, \cite{alemohammad2020recurrent} then derived the CK and NTK, known as the Recurrent NTK (RNTK), for RNNs, further show that classifiers equipped with RNTK outperform classical kernels, NTK of MLP, and finite RNNs on 56 time-series datasets from the UCR data repository \cite{Dau_2019}.


However, prior research focuses exclusively on simple form of RNNs \cite{ElmanRNN}, leaving out more complex RNN architectures such as bidirectional RNNs and RNNs with average pooling. 
Moreover, although overparametrized RNNs are universal approximators for any kind of data\cite{panigrahi2021learning}, they are almost exclusively used for time-series data applications such as language translation, speech transcription and audio processing \cite{2014arXiv1409.0473B,graves2013speech,sst2013}. Their utilities for non-time-series data have not been fully explored and demonstrated.
%
Last but not least, naive implementation of RNTK is computationally expensive because of quadratic computational complexity with respect to the number of training samples, which prevents RNTK's practical usage. Therefore, it is necessary to reduce and optimize the computational time for calculating the kernels associated with the recurrent architectures.

\noindent{\bf Contributions.} In this work, we focus on practical utility of infinite-width RNNs and its variants on non-time-series data and list our contributions as:  




[C1] Building upon previous work that derived the CK and NTK of multi-layer simple RNNs \cite{alemohammad2020recurrent}, we derive the CK and NTK of more complex RNN architectures including bi-directional RNN (BI-RNN), average pooling RNN (RNN-AVG) and their combination, BI-RNN-AVG. 

[C2] We provide the code to calculate the CK and NTK of RNNs and their variants when all data are of the same length. The code can be executed on CPU or GPU.

[C3] We show superior performance of infinite-width RNN-based kernels compared to other classical kernels and NTKs on a wide range of non time-series datasets. 


\section{Background}

We first review the key concepts of infinite-width DNNs, RNNs, and kernels associated with infinite-width RNNs, which will aid the development of the kernels for different RNN variants.

\textbf{Notations.}~ We denote $ [ n ] =  \{1, \dots, n\}$. $[\mA]_{i,j}$ represents the $(i,j)$-th entry of a matrix, and similarly $[\va]_i$ represents the $i$-th entry of a vector. $\boldsymbol{0}_d$ is a $d \times d$ matrix with all zero entries, and $\boldsymbol{I}_d$ is the identity matrix identity matrix of size $d$. 
We use $\phi(\cdot): \mathbb{R} \rightarrow \mathbb{R}$ to represent the activation function that acts coordinate wise on a vector and $\phi'$ to denote its derivative. 
We will use the rectified linear unit (ReLU) $\phi(x) = \mathrm{max}(0,x)$ in this paper.
$\mathcal{N}(\vmu, \mSigma)$ represents the multidimensional Gaussian distribution with the mean vector $\vmu$ and the covariance matrix $\mSigma$. We use $\vx\in\mathbb{R}^T$ to denote an input data point (vector) of dimension $T$.

\begin{figure*}[pt!]
    \centering
    \begin{minipage}{0.7\linewidth}
    \resizebox{\textwidth}{!}{%
\begin{tikzpicture}[object/.style={thin,double,<->}]
\tikzstyle{main}=[rectangle, minimum width = 16.3mm, minimum height = 7mm, thick, draw =black!80, node distance = 5mm]
\tikzstyle{main2}=[circle, minimum size = 10mm, thick, draw =black!80, node distance = 5mm]
\tikzstyle{connect}=[-latex, thick]
\tikzstyle{box}=[rectangle, draw=black!100]

  \node[main] (hlt) [] {\small$\vh^{(2,2)}(\vx)$};
  \node[main] (hltm) [left=1.3cm of hlt] {\small $\vh^{(2,1)}(\vx)$};
  \node[main] (hltp) [right=1.3cm of hlt] {\small $\vh^{(2,3)}(\vx)$};
  \node[main] (hlmt) [below=0.5cm of hlt] {\small $\vh^{(1,2)}(\vx)$};
  \node[main] (hlmtm) [below=0.5cm of hltm] {\small $\vh^{(1,1)}(\vx)$};
  \node[main] (hlmtp) [below=0.5cm of hltp] {\small $\vh^{(1,3)}(\vx)$};
  \coordinate[right=0.8cm of hltp] (hltpp);
  \coordinate[right=0.8cm of hlmtp] (hlmtpp);
  \coordinate[above=0.5cm of hltm] (hlptm);
  \coordinate[above=0.5cm of hlt] (hlpt);
  \coordinate[above=0.5cm of hltp] (hlptp);
  
  \path (hltm) edge [connect] node[above] {\small$\mW^{(2)}$}(hlt)
        (hlt) edge [connect] node[above] {\small$\mW^{(2)}$}(hltp)
        (hlmtm) edge [connect] node[above] {\small$\mW^{(1)}$}(hlmt)
        (hlmt) edge [connect] node[above] {\small$\mW^{(1)}$}(hlmtp)
		(hlmt) edge [connect] node[above] {\small$\mW^{(1)}$} (hlmtp)
		(hltp) edge [connect] (hltpp)
		(hlmtp) edge [connect] (hlmtpp)
		(hltm) edge [connect] (hlptm)
		(hlt) edge [connect] (hlpt)
		(hltp) edge [connect] (hlptp)
		(hlmtm) edge [connect] node[right] {\small$\mU^{(2)}$}(hltm)
		(hlmt) edge [connect] node[right] {\small$\mU^{(2)}$}(hlt)
		(hlmtp) edge [connect] node[right] {\small$\mU^{(2)}$} (hltp)
		;
	\node[rectangle, inner sep=.5mm,draw=black!100, fit= (hlt), fill=blue,opacity=.2](t) {};
	\node[rectangle, inner sep=1.7mm,draw=black!100, fit= (hltm)(hltp)(hltpp), fill=red,opacity=.2](t) {};

	
 \node[main] (hltz) [left=1.3cm of hltm] {\small $\vh^{(2,0)}$};
  \node[main] (hlmtz) [left=1.3cm of hlmtm] {\small $\vh^{(1,0)}$};
  \path (hltz) edge [connect] node[above] {\small$\mW^{(2)}$}(hltm)
        (hlmtz) edge [connect] node[above] {\small$\mW^{(1)}$}(hlmtm)
        ;

\node[rectangle, inner sep=1.mm,draw=black!100, fit= (hltz), fill=green,opacity=.2](t) {};
\node[rectangle, inner sep=1.mm,draw=black!100, fit= (hlmtz), fill=green,opacity=.2](t) {};
	

\node[main] (xtm) [below=0.5cm of hlmtm] {\small $[\vx]_{1}$};
\node[main] (xt)  [below=0.5cm of hlmt] {\small $[\vx]_{2}$};
\node[main] (xtp) [below=0.5cm of hlmtp] {\small $[\vx]_{3}$};
\path (xtm) edge [connect] node[right] {\small$\mU^{(1)}$}(hlmtm)
(xt) edge [connect] node[right] {\small$\mU^{(1)}$}(hlmt)
(xtp) edge [connect] node[right] {\small$\mU^{(1)}$}(hlmtp);

\end{tikzpicture}}

\end{minipage}
\begin{minipage}{0.24\linewidth}
    \caption{ \small
    Visualization of an RNN that highlights a cell (purple), a layer (red), and the initial hidden state of each layer (green). The particularity of RNN models lies in the recursion applied onto their input akin to infinite impulse response filters in signal processing. (Best viewed in color.) 
    \vspace{-3mm}
    }
    \label{fig:rnn}
    \end{minipage}
\end{figure*}
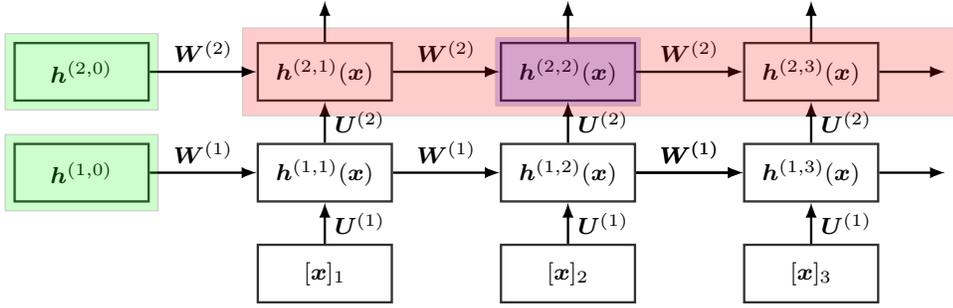

\subsection{Kernels for Infinite-Width DNNs}

Recent work has established a rigorous link between DNNs and kernel machines by studying {\em infinite-width regime} DNNs, in which the size of the hidden layers extends to infinity. For an infinite-width DNN $f_\theta(\vx)$ {\it at initialization}, i.e., whose parameters $\theta$ are initialized with independently and identically distributed (i.i.d) Gaussians, its output is drawn from a Gaussian Process (GP) with an associated Conjugate Kernel (CK) (also known as the NN-GP kernel) in the infinite-width limit \cite{neal1996priors}:
\begin{align} 
    \mathcal{K}(\vx,\vx') :=  \mathop{\E}_{\theta \sim \mathcal{N}}\big[ f_{\theta}(\vx) \boldsymbol{\cdot} f_{\theta}(\vx') \big], \label{NNGP}
\end{align}
where $\vx$ and $\vx'$ are two input data points. 

For an infinite-width DNNs that is {\it trained}, i.e., whose parameters $\theta$ are trained using gradient descent with respect to some loss function, prior works have shown that the DNN evolves as a linear model during training~\cite{lee2019wide}. More precisely, for parameters close to initialization, $\theta_0$, we can approximate $f_\theta(\vx)$ with the first-order Taylor expansion: 
\begin{align}
    f^{\rm lin}_{\theta}(\vx) \approx f_{\theta_0}(\vx) + \left\langle \nabla_{\theta}f_{\theta_0}(\vx) , \theta - \theta_0\right\rangle, \nonumber
\end{align} 
and we can use this linearized model for training $\theta$ instead of $f_\theta$. In this setting, the dynamics of training using gradient descent are equivalent to kernel gradient descent on the function space with respect to the NTK which is  formulated as 
\begin{align}
    \Theta(\vx,\vx') := \left\langle \nabla_{\theta}f_{\theta}(\vx) , \nabla_{\theta}f_{\theta}(\vx')\right\rangle. \label{NTK}
\end{align}
for any DNN architecture \cite{jacot2018neural, yang2021tensor}. 

\subsection{Simple Multi-Layer RNNs for Non-Time-Series Data}

To denote non-time-series data $\vx \in \mathbb{R}^T$ for RNNs, we will use $[\vx]_t$ as the data fed to the RNN at time $t$. Hence, a RNN with $n$ units in each hidden layer $\ell$ performs the following recursive computation for $t \in [T]$ and $\ell \in [L] $
\begin{align} 
    \vg^{\left(1,t\right)}(\vx) &= \frac{\sigma_w}{\sqrt{n}}\rmW^{\left(1\right)}\vh^{\left(1,t-1\right)}(\vx) + \sigma_u \rmU^{\left(1\right)}[\vx]_t + \sigma_b\rvb^{\left(1\right)}, \nonumber \\
	\vg^{\left(\ell,t\right)}(\vx) &= \frac{\sigma_w}{\sqrt{n}}\rmW^{\left(\ell\right)}\vh^{\left(\ell,t-1\right)}(\vx) + \frac{\sigma_u}{\sqrt{n}}\rmU^{\left(\ell\right)}\vh^{\left(\ell-1,t\right)}(\vx) + \sigma_b\rvb^{\left(\ell\right)}, \nonumber \\
	\vh^{\left(\ell,t\right)}(\vx) &= \phi\left(  \vg^{\left(\ell,t\right)}(\vx) \right).\, 
	\nonumber
\end{align}
Where $\mW^{(\ell)} \in \mathbb{R}^{n \times n}$, $ \vb^{(\ell)} \in \mathbb{R}^{n}$ for any $\ell \in [L]$ , $\mU^{(1)} \in \mathbb{R}^{1\times n} $ and $ \mU^{(\ell)} \in \mathbb{R}^{n \times n}$ for $\ell \ge 2$. We set the initial hidden state in all layers to zero, i.e, $	\vh^{\left(\ell,0\right)} = 0$. 
The output of an RNN is typically a linear transformation of the hidden layer at the last layer $L$ and at each time $t$: 
\begin{align}
    &f^{(t)}_\theta(\vx) = \frac{\sigma_v}{\sqrt{n}}\rmV^{(t)} \vh^{(L,t)}(\vx) \in \mathbb{R}^d. \label{eq:rnn-last-layer}
\end{align} 
For simple RNNs, we keep only the output at the last time step by replacing $t$ in equation~(\ref{eq:rnn-last-layer}) to $T$. The learnable parameters 
\begin{align} 
    \theta = \mathrm{vect}\big[\{\{\rmW^{(\ell)},\rmU^{(\ell)},\rvb^{(\ell)}\}_{\ell = 1}^{L} , \rmV^{(T)} \} \big], 
    \label{params}
\end{align}
are initialized with a standard Gaussian, i.e., $\mathcal{N}(0,1)$ and
$\sigma = \{\sigma_w,\sigma_u,\sigma_b,\sigma_v \}$ 
are initialization hyperparameters that control the scale of the parameters.

Fig.~\ref{fig:rnn} visualizes an RNN. 
Note that, because we focus on {\it non time-series} data in this work, $T$ is fixed for all data points in a given dataset. Therefore, we refer to $t$ as the $t$-th dimension in a data point $\vx$ instead of the $t$-th time step.

\subsection{CK and NTK of  Infinite-Width simple RNNs}

Because of the equivalence between infinite-width DNNs and GPs \cite{lee2017deep, duvenaud2014avoiding,novak2018bayesian, garrigaalonso2018deep,yang2019tensor}, as $n \rightarrow \infty$, each coordinate of the RNN pre-activation $\vg^{\left(\ell,t\right)}(\vx)$ vectors also converges to zero mean GPs with the kernel: 
\begin{align} 
    \Sigma^{(\ell,t)}(\vx,\vx') &= \mathop{\E}_{\theta \sim \mathcal{N}} \big[ [\vg^{(\ell,t)}(\vx)]_i \boldsymbol{\cdot} [\vg^{(\ell,t)}{(\vx')}]_i  \big]\, \,\, \forall i \in [n]. \label{eq:pre-ggp}
\end{align}
Consequently, each coordinate of a simple RNN output converges to a GP with the CK kernel
\begin{align} 
    \mathcal{K}^{(T)}(\vx,\vx') & =  \mathop{\E}_{\theta \sim \mathcal{N}}\big[ [f^{(T)}_{\theta}(\vx)]_i \boldsymbol{\cdot} [f^{(T)}_{\theta}(\vx')]_i \big], \hspace{1mm} \forall i \in [d]. 
\end{align}
Also, for the gradient vectors $\vdelta^{(\ell,t)}(\vx) := \sqrt{n} \big( \nabla_{\vg^{(\ell,t)}(\vx)} f_{\theta}(\vx) \big)$ we have
\begin{align} 
    \Pi^{(\ell,t)}(\vx,\vx') &= \mathop{\E}_{\theta \sim \mathcal{N}} \big[ [\vdelta^{(\ell,t)}(\vx)]_i \boldsymbol{\cdot} [\vdelta^{(\ell,t)}(\vx')]_i  \big]\, \,\, \forall i \in [n]. \label{eq:grad-gp}
\end{align}
The NTK of a simple RNN for two inputs $\vx, \vx' \in \mathbb{R}^T$ is  obtained as
\begin{align}
    &\Theta(\vx,\vx') = \left\langle \nabla_{\theta}f^{(T)}_{\theta}(\vx) , \nabla_{\theta}f^{(T)}_{\theta}(\vx')\right\rangle = \nonumber\\ &\left( \sum_{\ell =  1}^{L} \sum_{t = 1}^{T} \left( \Pi^{(\ell,t)}(\vx,\vx') \boldsymbol{\cdot} \Sigma^{(\ell,t)}(\vx,\vx') \right) +  \mathcal{K}^{(T)}(\vx,\vx') \right)  \otimes \mI_d ,\label{eq:rntk} 
\end{align}

For more details on the derivation of the CK and NTK of the simple RNN see \cite{alemohammad2020recurrent}.
The above CK and NTK kernels associated with simple RNNs in their infinite width regime lays the ground work for our derivation for the CK and NTK kernels associated with various other RNN architectures.

\section{Infinite-Width Bidirectional And Average Pooling RNN Kernels}

In this section, we derive the CK and NTK of different variants of RNNs: bidirectional RNN, RNN with average pooling, and their combination. Our derivations are based on the formula for the CK and NTK of the single output in the last time step in equations (\ref{eq:pre-ggp}), (\ref{eq:grad-gp}), (\ref{eq:rntk}) and the following lemma introduced in \cite{yang2019scaling}.

\begin{lm} 
\label{lm:1}
let $\vh(\vx)$ and $\Bar{\vh}(\vx')$ be two arbitrary hidden states in an infinite-width DNN with parameters $\theta$. Take two outputs $y(\vx) = \frac{\sigma_v}{\sqrt{n}}\rmV \vh(\vx)$ and $\Bar{y}(\vx') = \frac{\sigma_v}{\sqrt{n}}\Bar{\rmV}\Bar{\vh}(\vx')$, where $\rmV,\Bar{\rmV} \in \mathbb{R}^{n \times d}$ are drawn \emph{independently} from the Gaussian distribution. The following quantities are always zero regardless of the DNN architecture,
\begin{align}
  \mathop{\E}_{\theta \sim \mathcal{N}}\big[ y_i(\vx) \times \Bar{y}_j(\vx')^T \big] &= \boldsymbol{0}_d \nonumber\\
 \left\langle \nabla_{\theta}y(\vx) ,  \nabla_{\theta}\Bar{y}(\vx')\right\rangle &= \boldsymbol{0}_d.  \nonumber
\end{align}
\end{lm}

\subsection{RNNs with Average Pooling}

Average pooling is defined as follows
\begin{align}
    f_{\theta}^{\rm avg}(\vx) = \sum_{t = 1}^{T} \frac{\sigma_v}{\sqrt{n}}\rmV^{(t)} \vh^{(L,t)}(\vx) = \sum_{t = 1}^{T}  f^{(t)}_\theta(\vx). \label{avg}
\end{align}

Where we use different weights for the output. In this case, the data points must be the same dimension; therefore, we have a fixed architecture. $\theta = \mathrm{vect}\big[\{\{\rmW^{(\ell)},\rmU^{(\ell)},\rvb^{(\ell)}\}_{\ell = 1}^{L} , \{ \rmV^{(t)} \}_{t=1}^{T} \} \big]$ forms the fixed parameters set. Hence, the NTK becomes
\begin{align}
    \Theta^{\rm avg}(\vx,\vx') &= \left\langle \nabla_{\theta}f^{\rm avg}_{\theta}(\vx) , \nabla_{\theta}f^{\rm avg}_{\theta}(\vx')\right\rangle \nonumber\\ &= \sum_{t = 1}^{T}  \sum_{t' = 1}^{T} \left\langle \nabla_{\theta}f^{(t)}_{\theta}(\vx),  \nabla_{\theta}f^{(t')}_{\theta}(\vx). \right\rangle \nonumber
\end{align}
Because of independence of the output layer weights for each time step, based on Lemma \ref{lm:1} we have 
\begin{align}
    \left\langle \nabla_{\theta}f^{(t)}_{\theta}(\vx),  \nabla_{\theta}f^{(t')}_{\theta}(\vx) \right\rangle = 0 \hspace{2.5mm} t \ne t' \label{32}
\end{align}
As a result
\begin{align}
    \Theta^{\rm avg}(\vx,\vx')  &= \sum_{t = 1}^{T} \left\langle \nabla_{\theta}f^{(t)}_{\theta}(\vx),  \nabla_{\theta}f^{(t')}_{\theta}(\vx) \right\rangle = \sum_{t = 1}^{T} \Theta^{(t)}(\vx,\vx').\nonumber
\end{align}
Similarly, for the CK we have
\begin{align}
    \mathcal{K}^{\rm avg}(\vx,\vx') &= \sum_{t = 1}^{T} \mathop{\E}_{\theta \sim \mathcal{N}}\big[ [f^{(t)}_{\theta}(\vx)]_i \boldsymbol{\cdot} [f^{t}_{\theta}(\vx')]_i \big] = \sum_{t = 1}^{T} \mathcal{K}^{(t)}(\vx,\vx').\nonumber
\end{align}
Each $ \Theta^{(t)}(\vx,\vx')$ and $\mathcal{K}^{(t)}(\vx,\vx')$ can be thought of as the CK and NTK kernels of an RNN with a single output at last time step when the first $t$ data points are fed to the network.

\textbf{Remark.} An alternative approach to average pooling uses the same weights for the output at each time step, i.e.,
\begin{align}
    f_{\theta}^{\rm avg}(\vx) = \sum_{t = 1}^{T} \frac{\sigma_v}{\sqrt{n}}\rmV\vh^{(L,t)}(\vx), \nonumber
\end{align}
which is necessary to handle signals of different lengths. Here, the outputs between different time steps will be correlated, because equation (\ref{32}) will have a nonzero value. However, calculating the NTK with this pooling strategy requires storing a tensor of dimension $(T \times T \times N \times N)$, where $N$ is the training data size and $T$ the data length. While block computation and approximations could be developed for such cases, we instead concentrate on the pooling strategy introduced in \ref{avg} as it provides  an out-of-the-box tractable implementation. For further details, see \cite{ yang2020tensor} on how the NTK of a single layer RNN with average pooling using the same output weight is calculated.

\begin{algorithm}[t]
\caption{CK of RNN and RNN-AVG}
\label{alg:gpkernels}
\begin{algorithmic}[1]
\For{$t=1$}
    \State $\Sigma^{(1,1)}(\vx,\vx') = \sigma_u^2 ([\vx]_1\boldsymbol{\cdot}[\vx']_1) + \sigma_b^2$
    \For{$\ell=2,\dots,L$}
        \State $\Sigma^{(\ell,1)}(\vx,\vx') = \sigma_u^2 \mathrm{V}_{\phi}\big[\mK^{(\ell-1,1)}(\vx,\vx') \big] + \sigma_b^2$
    \EndFor
    \State $ \mathcal{K}^{(1)}(\vx,\vx') = \sigma_v^2 \mathrm{V}_{\phi}\big[\mK^{(L,1)}(\vx,\vx') \big] $
    \State $\mathcal{K}^{\rm avg}(\vx,\vx') = \mathcal{K}^{(1)}(\vx,\vx')$
\EndFor
\For{$t=2,\dots,T$}
    \State $\Sigma^{(1,t)}(\vx,\vx') =  \sigma_u^2 ([\vx]_t\boldsymbol{\cdot}[\vx']_t) + \sigma_w^2\mathrm{V}_{\phi}\big[\mK^{(1,t-1)}(\vx,\vx') \big]$
    \Statex $ \hspace{2.55cm} + \sigma_b^2$
    \For{$\ell=2,\dots,L$}
        \State $\Sigma^{(\ell,t)}(\vx,\vx') = \sigma_u^2\mathrm{V}_{\phi}\big[\mK^{(\ell-1,t)}(\vx,\vx') \big]$
        \Statex \hspace{2.9cm} $
        + \hspace{0.5mm} \sigma_w^2\mathrm{V}_{\phi}\big[\mK^{(\ell,t-1)}(\vx,\vx') \big] + \sigma_b^2$
    \EndFor
    \State $ \mathcal{K}^{(t)}(\vx,\vx') = \sigma_v^2 \mathrm{V}_{\phi}\big[\mK^{(L,t)}(\vx,\vx') \big] $
    \State $\mathcal{K}^{\rm avg}(\vx,\vx') \leftarrow \mathcal{K}^{\rm avg}(\vx,\vx') + \mathcal{K}^{(t)}(\vx,\vx') $
\EndFor
\State  \textbf{CK of RNN:} $\mathcal{K}^{(T)}(\vx,\vx')$
\State \textbf{CK of RNN-AVG:} $\mathcal{K}^{\rm avg}(\vx,\vx')$
\end{algorithmic}
\end{algorithm}

\begin{algorithm}[t]
\caption{NTK of RNN-AVG}
\label{alg:rntk}
\begin{algorithmic}[1]
\For{$t=1$}
    \State $\Psi^{(1,1)}(\vx,\vx') = \Sigma^{(1,1)}(\vx,\vx')$
    \For{$\ell=2,\dots,L$}
        \State $\Psi^{(\ell,1)}(\vx,\vx') = \Sigma^{(\ell,1)}(\vx,\vx')$ 
        \Statex \hspace{2.75 cm}$ + \hspace{0.5mm}  \sigma_u^2\Psi^{(\ell-1,1)}(\vx,\vx')\mathrm{V}_{\phi'}\big[\mK^{(\ell-1,1)}(\vx,\vx') \big]$
    \EndFor
    \State $\Theta^{(1)}(\vx,\vx') = \mathcal{K}^{(1)}(\vx,\vx')$
    \Statex \hspace{1.98 cm} $ +  \sigma_v^2\Psi^{(L,1)}(\vx,\vx')\mathrm{V}_{\phi'}\hspace{-0.061cm}\big[\mK^{(L,1)}(\vx,\vx') \big]$
    \State $\Theta^{\rm avg}(\vx,\vx') = \Theta^{(1)}(\vx,\vx')$
\EndFor
\For{$t=2,\dots,T$}
    \State $\Psi^{(1,t)}(\vx,\vx') = \Sigma^{(1,t)}(\vx,\vx')$
    \Statex \hspace{2.15 cm} $ + \hspace{0.5mm}  \sigma_w^2\Psi^{(1,t-1)}(\vx,\vx')\mathrm{V}_{\phi'}\big[\mK^{(1,t-1)}(\vx,\vx') \big]$
    \For{$\ell=2,\dots,L$}
        \State $\Psi^{(\ell,t)}(\vx,\vx') = \Sigma^{(\ell,t)}(\vx,\vx)$
        \Statex \hspace{2.7 cm}$ + \hspace{0.5mm}  \sigma_w^2\Psi^{(\ell,t-1)}(\vx,\vx')\mathrm{V}_{\phi'}\big[\mK^{(\ell,t-1)}(\vx,\vx') \big]$
        \Statex \hspace{2.7 cm}$ + \hspace{0.85mm}  \sigma_u^2\Psi^{(\ell-1,t)}(\vx,\vx')\mathrm{V}_{\phi'}\big[\mK^{(\ell-1,t)}(\vx,\vx') \big]$
    \EndFor
     \State $\Theta^{(t)}(\vx,\vx') = \mathcal{K}^{(t)}(\vx,\vx')$
    \Statex \hspace{1.95 cm} $+  \sigma_v^2\Psi^{(L,t)}(\vx,\vx')\mathrm{V}_{\phi'}\hspace{-0.061cm}\big[\mK^{(L,t)}(\vx,\vx') \big]$
    \State $\Theta^{\rm avg}(\vx,\vx') \leftarrow \Theta^{\rm avg}(\vx,\vx') + \Theta^{(t)}(\vx,\vx')$
\EndFor
\State \textbf{NTK of RNN:} $\Theta^{(T)}(\vx,\vx')$
\State \textbf{NTK of RNN-AVG:} $\Theta^{\rm avg}(\vx,\vx')$
\end{algorithmic}
\end{algorithm}

\subsection{Infinite-Width RNN Kernels Pseudo Code}

\begin{figure*}[t]
    \centering 
\begin{minipage}{0.01\textwidth}
\rotatebox{90}{\;\;\; {\scriptsize Computation Time (seconds)}}
\end{minipage}
\begin{minipage}{0.325\textwidth}
    \centering
    \subfloat{\includegraphics[width=\linewidth]{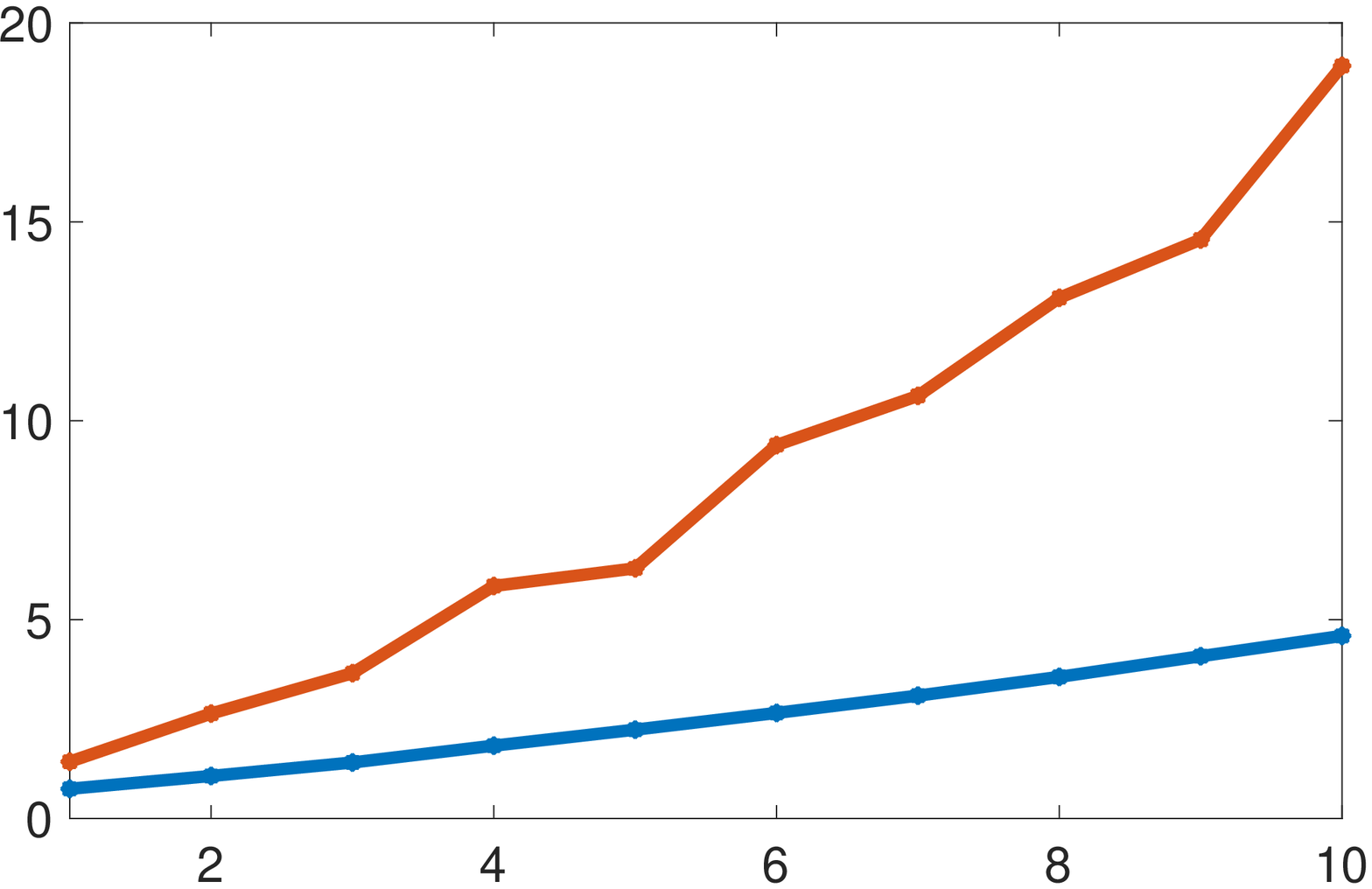} \label{fig2a} }\\
    {\vspace{0.01mm} \small\hspace{1mm} $L$} 
\end{minipage}\hfil 
\begin{minipage}{0.325\textwidth}
    \centering
    \subfloat{\includegraphics[width=\linewidth]{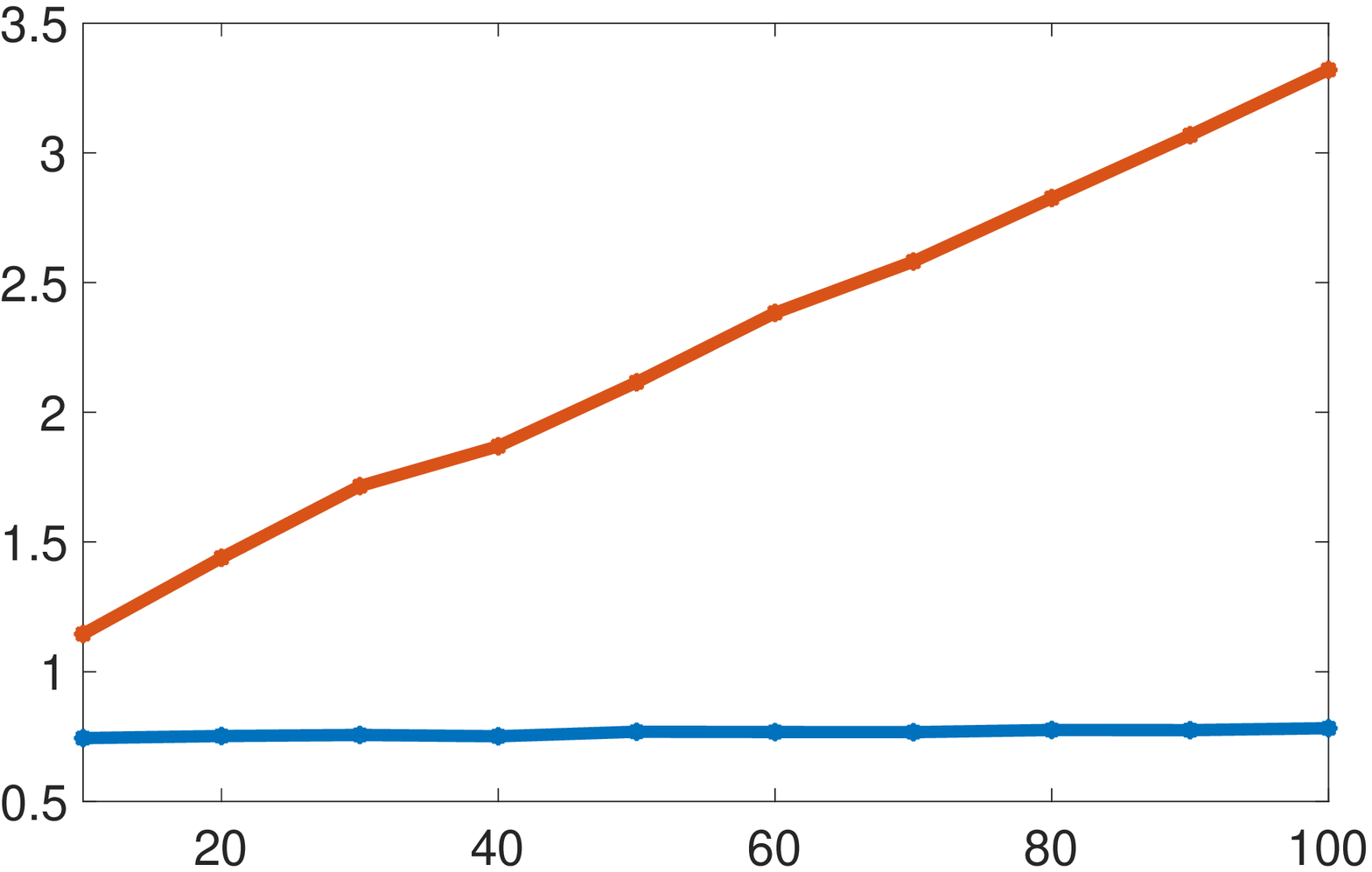} \label{fig2b}}\\
    {\vspace{0.01mm} \small \hspace{1mm} $T$} 
\end{minipage}\hfil 
\begin{minipage}{0.325\textwidth}
    \centering
    \subfloat{\includegraphics[width=\linewidth]{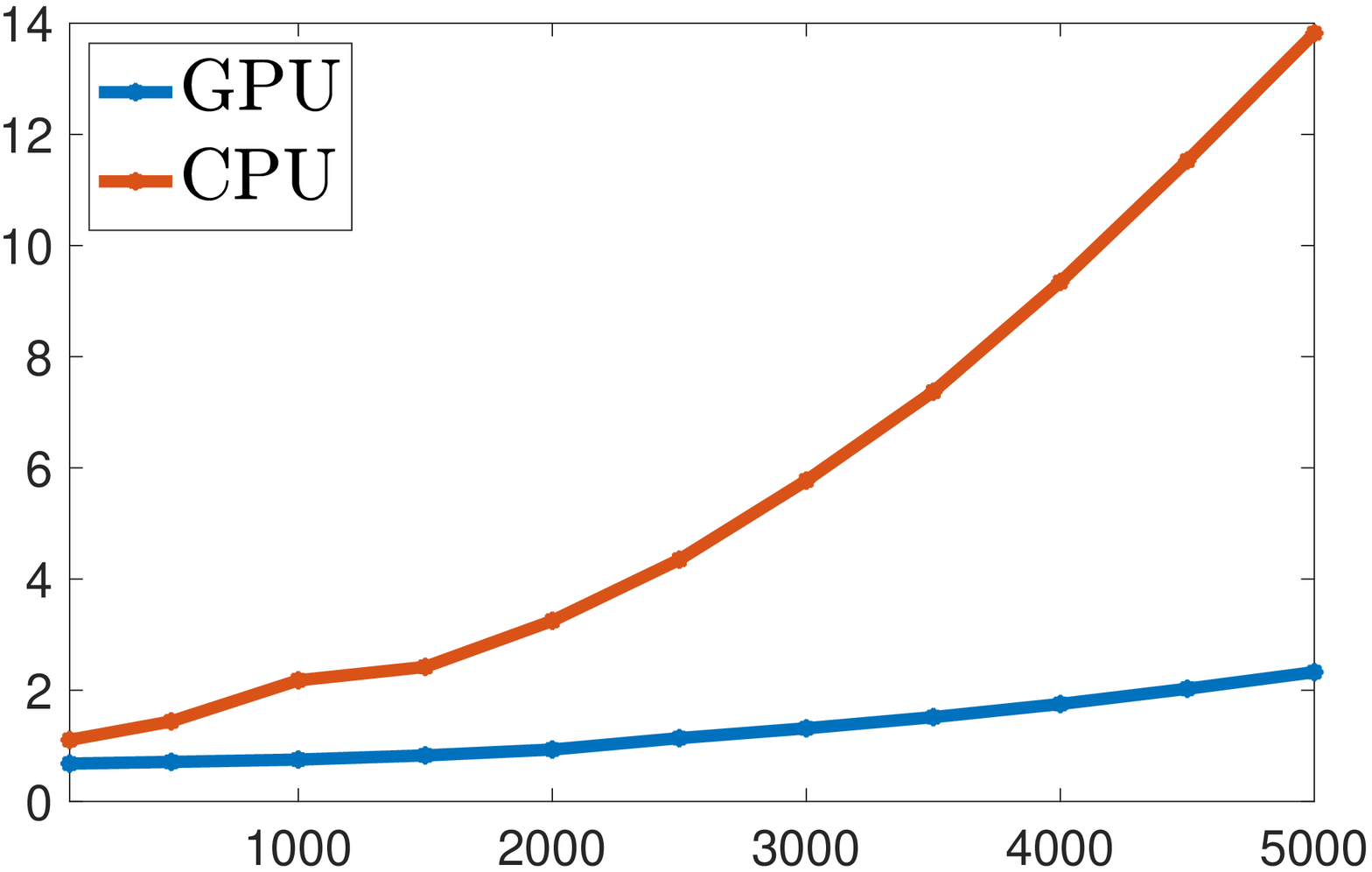} \label{fig2c}}\\ 
    {\vspace{0.01mm} \small \hspace{1mm} $N$} 
\end{minipage}
\caption{\small
Average computation time over $100$ artificial random datasets for GPU (\textbf{blue}) and CPU (\textbf{red}) for different number of RNTK layers ($L$), data length ($T$) and number of data points ($N$). \textbf{Left}: $T=20$, $N=1000$ and vary $L$ \textbf{Middle}: $L=1$, $N=1000$ and vary $T$. \textbf{Right}: $L=1$, $T=20$ and vary $N$. Through these three experiments we can see how the GPU implementation significantly reduces the computation time compared to CPU implementation with increasing $L$, $T$ and $N$.
}
\label{fig:cpu_gpu}
\end{figure*}

We provide in Algo.~\ref{alg:rntk} the pseudo code of the derived kernels of the RNN and RNN-AVG architectures. We also provide a CPU and GPU implementation for which we report various computation times in Fig.~\ref{fig:cpu_gpu} to demonstrate how the proposed kernels are scalable on GPU. We can for example observe how the data-length has little effect on the computational time (linear trend on CPU, constant on GPU). The same goes for the depth of the employed RNN architecture. Lastly, the number of samples also produces a near linear trend when the GPU implementation is employed, on CPU, the expected quadratic trend can be observed. Those observations hold as long as the combination of depth/dataset size/data-length allow the computations to fit in GPU memory.

\subsection{Bidirectional RNNs}

In BI-RNNs, the original signal $\vx$ is fed into a simple RNN with parameters $\theta$ in 
equation (\ref{params}) and hyperparameters $\sigma$ to calculate the hidden states $\vh^{(\ell,t)}(\vx)$ and the output $f^{(T)}_{\theta}(\vx)$. In addition, the flipped version of the signal $\vx$, i.e $\Bar{\vx} = \{ \vx_{T-t} \}_{t = 0}^{T-1}$, is fed to another simple RNN structure with the same initialization hyperparameters $\sigma$, but parameters $\ddot{\theta}$ are iid copies of $\theta$ to produce $\ddot{\vh}^{(\ell,t)}(\Bar{\vx})$ and $\ddot{f}^{(T)}_{\ddot{\theta}}(\Bar{\vx})$. The output of a BI-RNN is simply the sum of the output of two networks
\begin{align}
    f_{\Tilde{\theta}}^{\rm bi}(\vx) = f^{(T)}_{\theta}(\vx) + \ddot{f}^{(T)}_{\ddot{\theta}}(\Bar{\vx}),\nonumber
\end{align}
with $\Tilde{\theta} = \theta \cup \ddot{\theta}$ is the set of all parameters. The NTK of this network becomes
\begin{align}
     \Theta^{\rm avg}(\vx,\vx') = &\langle \nabla_{\Tilde{\theta}}f^{\rm bi}_{\Tilde{\theta}}(\vx) , \nabla_{\Tilde{\theta}}f^{\rm bi}_{\Tilde{\theta}}(\vx')\rangle \nonumber\\ = &\langle \nabla_{\theta}f^{(T)}_{\theta}(\vx) , \nabla_{\theta}f^{(T)}_{\theta}(\vx')\rangle \nonumber \\ &+ \langle \nabla_{\ddot{\theta}}f^{(T)}_{\ddot{\theta}}(\Bar{\vx}) , \nabla_{\ddot{\theta}}f^{(T)}_{\ddot{\theta}}(\Bar{\vx}')\rangle \nonumber\\ &+ \langle \nabla_{\ddot{\theta}}f^{(T)}_{\ddot{\theta}}(\Bar{\vx}) , \nabla_{\theta}f^{(T)}_{\theta}(\vx')\rangle \nonumber \\ &+ \langle \nabla_{\theta}f^{(T)}_{\theta}(\vx) , \nabla_{\ddot{\theta}}f^{(T)}_{\ddot{\theta}}(\Bar{\vx}')\rangle, \label{42}
\end{align}
where the third and the fourth terms in equation (\ref{42}) become zero as a result of Lemma \ref{lm:1}. Thus, we have:
\begin{align}
     \Theta^{\rm bi}(\vx,\vx') = \Theta(\vx,\vx') + \Theta(\Bar{\vx},\Bar{\vx}'), \label{CKBIRNN} 
\end{align}
and 
\begin{align}
    \mathcal{K}^{\rm bi}(\vx,\vx') =  \mathcal{K}^{(T)}(\vx,\vx') +  \mathcal{K}^{(T)}(\Bar{\vx},\Bar{\vx}'). \label{NTKBIRNN} 
\end{align}
\subsection{Bidirectional RNNs with Average Pooling}
Calculation of the CK and NTK of a BI-RNN-AVG follows the same procedure as a BI-RNN. The kernels will be the sum of kernels of RNN-AVG evaluated on both versions of input data 
\begin{align}
    \Theta^{\rm bi-avg}(\vx,\vx') &= \Theta^{\rm avg}(\vx,\vx') + \Theta^{\rm avg}(\Bar{\vx},\Bar{\vx}'), \label{NTKBIAVGRNN} \\
    \mathcal{K}^{\rm bi-avg}(\vx,\vx') &=  \mathcal{K}^{\rm avg}(\vx,\vx') +  \mathcal{K}^{\rm avg}(\Bar{\vx},\Bar{\vx}'). \label{CKBIAVGRNN}
\end{align}

\section{Fast GPU-Based Implementation} \label{gpufast}

In this section, we provide the implementation details of calculating the CK and NTK Gram matrices associated with infinite-width RNNs of the same length. We also demonstrate the running time of calculating the Gram matrices on a CPU and GPU.

Generally, kernels associated with RNNs can handle data of various lengths \cite{alemohammad2020recurrent}. In case two input sequences of data have different lengths, the calculation of the CK and NTK requires an adaptive implementation with respect to the difference of the length of inputs, which impedes efficient parallelization of the computation of the Gram matrix with currently available toolboxes. However, when all the data are of the same length, the pairwise calculation of CK and NTK of two inputs follows a combined procedure. In this setting, the kernels for all pairwise data can be computed at once using simple matrix computations. Such computations can be highly parallelized on GPU, enabling the fast application of kernels associated with infinite-width RNNs for kernel-based classification or regression tasks using the SymJAX library \cite{r2020symjax}.

Implementations details for the CK and NTK of RNN and RNN-AVG are provided in Algorithms \ref{alg:gpkernels} and \ref{alg:rntk}, respectively. The kernels of BI-RNN and BI-RNN-AVG can be obtained easily by summing the outputs of the algorithms with the original signal and with the flipped version as inputs as demonstrated in equations (\ref{CKBIRNN}), (\ref{NTKBIRNN}), (\ref{CKBIAVGRNN}), and (\ref{NTKBIAVGRNN}).

One important practical advantage of the Algorithms \ref{alg:gpkernels} and \ref{alg:rntk} is the memory efficiency. To obtain NTK of RNNs, one needs to calculate and store the GP kernels of pre-activation layers (equation (\ref{eq:pre-ggp}) ) and then gradient layers (equation (\ref{eq:grad-gp})) for all time steps, and then calculate NTK of RNNs using equation (\ref{eq:rntk}). However, with the assumption of the same length, the calculation of the NTK can be done simultaneously with the calculation of the GP without directly calculating the GP kernel of gradients in equation (\ref{eq:grad-gp}) using the Algorithm \ref{alg:rntk}. The implementation in Algorithm \ref{alg:rntk} for the NTK only needs GP kernels of the previous and current time step calculated at Algorithm \ref{alg:gpkernels}, as opposed to equation (\ref{eq:rntk}) which needs the GP kernels of all time steps. As a result, storage space of order $\mathcal{O}(N\times N)$ is needed to calculate the NTK, rather than $\mathcal{O}(T\times N \times N)$, which is crucial for data of long lengths.

Algorithms \ref{alg:gpkernels} and \ref{alg:rntk} depend on an operator $\mathrm{V}_{\phi}\big[\mK \big]$ that depends on the nonlinearity $\phi(\cdot)$ and a positive semi-definite matrix $\mK\in\mathbb{R}^{2 \times 2}$
\begin{align} 
    \mathrm{V}_{\phi}\big[\mK \big] = \E [\phi(\rz_1)\boldsymbol{\cdot}\phi(\rz_2)], \qquad (\rz_1,\rz_2) \sim \mathcal{N}(0,\mK). \nonumber
\end{align}

In the case of $\phi = \rm ReLU$, an analytical formula for $\mathrm{V}_{\phi}\big[\mK \big]$ and $\mathrm{V}_{\phi'}\big[\mK \big]$ exists \cite{cho2009kernel}. For any positive definite matrix $ \mK = \left[ {\begin{array}{cc} K_1 & K_3 \\ K_3 & K_2 \\ \end{array} } \right] $ we have
\begin{align}
    \mathrm{V}_{\phi}[\mK] &= \frac{1}{2\pi}\left(c(\pi - \rm arccos (c)) +\sqrt{1- c^2} ) \right)\sqrt{K_1K_2}, \nonumber \\
        \mathrm{V}_{\phi'}[\mK] &= \frac{1}{2\pi}(\pi - \rm arccos (c)), \nonumber
\end{align}
where $c = K_3/\sqrt{K_1K_2}$.

To compute the RNN kernels in this paper, the $\mK^{(\ell,t)}(\vx,\vx')$ is defined as
\begin{align}
        \mK^{(\ell,t)}(\vx,\vx') &=  \left[ {\begin{array}{cc} \Sigma^{(\ell,t)}(\vx,\vx) & \Sigma^{(\ell,t)}(\vx,\vx') \\ \Sigma^{(\ell ,t)}(\vx,\vx') & \Sigma^{(\ell,t)}(\vx',\vx') \\ \end{array} } \right]. \nonumber 
\end{align}

\begin{table*}
\centering
\caption{Summary of non time-series classification results on 90 UCI datasets. RNN-P outperforms all other methods.}
{
\begin{adjustbox}{width=\linewidth,center}
\begin{tabular}{|l|cccccccccc|}
\toprule
\textbf{}         & \textbf{RF} & \textbf{RBF} &\textbf{Polynomial} & \textbf{MLP} & \textbf{H-\textbf{$\gamma$}-exp} & \textbf{RNN} & \textbf{BI-RNN} & \textbf{RNN-AVG} & \textbf{BI-RNN-AVG} & \textbf{RNN-P}\\ \hline
Acc. mean $\uparrow$                & 81.56\%       & 81.03\%      & 80.91\%       & 81.95\%    & 82.25\%     & 81.73\%       & 81.94\%     & 81.75\%     & 82.07\%    & {\bf 82.34}\%    \\
Acc. std                 & 13.90\%       & 15.09\%     & 14.10\%       & 14.10\%    & 14.07\%    & 14.24\%       & 14.47\%      & 14.43\%     & 14.45\%    & 14.06\% \\
P95 $\uparrow$                     & 81.11\%       & 81.11\%     & 70.00\%       & 84.44\%    & {\bf 87.78}\%    & 83.33\%       & 84.44\%      & 82.22\%     & {\bf 87.78}\%    & 86.67\% \\
PMA $\uparrow$                     & 96.88\%       & 96.09\%     & 96.13\%       & 97.33\%    & 97.71\%    & 97.00\%       & 97.22\%      & 96.97\%     & 97.38\%    & {\bf 97.80}\% \\
Friedman Rank $\downarrow$            & 5.58           & 5.60        & 6.27          & 4.83       & 4.30        & 5.16           & 4.92        & 5.16         & 4.28        & {\bf 4.22}  \\
\bottomrule
\end{tabular}
\label{table:1}
\end{adjustbox}
}
\label{tab:real-data-svm}
\end{table*}

\subsection{CPU And GPU Computation Times}

In this section we present a detailed comparison between the CPU and GPU speed of the proposed NTK implementation. More specifically, we use an artificial dataset in order to control the number of data points, the dimension and length of the samples. 

In Figure.~\ref{fig:cpu_gpu} we provide the CPU and GPU computation times of Algorithm ~\ref{alg:rntk} for varying dataset parameters. For the CPU computation, we use a $10$-core Intel CPU i9-9820X with $62$G of RAM. For the GPU configuration, we used a single Nvidia GTX2080Ti card. In both cases the hardware was installed on a desktop computer.

One computational draw back for Algorithms \ref{alg:gpkernels} and \ref{alg:rntk} is linear complexity with respect to the dimension of data $T$ and number of layers $L$. Our GPU implementation is highly efficient with respect to $T$, which facilitates the computations for high dimensional data. All in all, the GPU implementation sees linear or sub-linear computation time increase when increasing any of those parameters while CPU implementation sees from exponential to linear computation time increases. 

We provide all implementations on Github in the following link \texttt{\url{https://github.com/SinaAlemohammad/RNTK_UCI}}. Implementations are done in SymJAX \cite{r2020symjax} that benefits from a highly optimized XLA backend compilation that provides optimized CPU or GPU executions. 

\section{Experiments}
\label{experiments}

In this section, we empirically validate the performance of C-SVM classifier associated with kernels for each RNN variant developed in the previous sections. In our experiments, we use the same datasets used in \cite{Arora2020Harnessing}, that include  $90$ non-time-series UCI pre-processed datasets from UCI data repository \cite{FERNANDEZDELGADO201911} (all the datasets from the full collection that have less than $5000$ samples).

\textbf{Performance and model comparison.}~We compare our methods  with the top classifiers evaluated in \cite{JMLR:v15:delgado14a}, which are random forests (RF), polynomial and Gaussian kernel SVM. In addition to those conventional methods, we also compare against the recently derived MLP NTK with C-SVM \cite{Arora2020Harnessing} and against a combination of Laplace and exponential kernels known as H-$\gamma$-exp \cite{geifman2020similarity} that delivers the best performance on those datasets among the previous mentioned techniques. 

\textbf{Testing procedure.}~The training procedure follows that of \cite{Arora2020Harnessing} and \cite{JMLR:v15:delgado14a}. Each dataset is divided into two subsets of equal size; one is used for training and the other for validation. After cross-validation has been completed and the best hyper-parameters have been obtained based on the validation set performance, the test accuracy is computed by averaging the accuracy of $K$-fold cross testing with $K=4$. That is, the dataset is split into $4$ folds, $3$ of which are used for training, based on the validation set's best hyperparameter, and the test accuracy is obtained from the remaining fold. This process is repeated for all $4$ folds. The average accuracy of those $4$ test sets is denoted as the overall model test accuracy. We use the same splitting indices presented in \cite{JMLR:v15:delgado14a} in our experiments.

For our model, the hyperparameters that are cross-validated as per the above procedure are the following
\begin{align}
    \sigma_u &\in \{ 0.25, 0.5 \} \nonumber\\
    \sigma_b &\in \{ 0.001, 0.1 \} \nonumber\\
    L & \in \{ 1,2 \}. \nonumber
\end{align}
As proposed in \cite{alemohammad2020recurrent}, we set $\sigma_w = \sqrt{2}$ in order for the kernel to consider all observations from different time-steps equally. Because $\sigma_v = 1$ merely scales the output of the a selected kernel,  this parameter can be adjusted specifically to allow the final pairwise kernel output to have similar scaling with respect to the C-SVM cost for different type of kernels such as BI-RNN, AVG-RNN, etc. Therefore, we use $\sigma_v = 1$ for simple RNN, $ \sigma_v = \frac{1}{ \sqrt{2} } $ for BI-RNN, $ \sigma_v = \frac{1}{\sqrt{T}}$ for AVG-RNN, and $\sigma_v = \frac{1}{\sqrt{2T}}$ for BI-RNN-AVG. 

Because we chose C-SVM as our classification method, we have an additional set of hyper-parameters to cross-validate
\begin{align}
    C \in \{ 0.01,1,100,10000,1000000\}. \nonumber
\end{align}
Those sets of hyperparameters are cross-validated independently for each dataset.

One remaining hyperparameter to consider is the sequence ordering of the input data. In general, in order to learn the mapping function, RNNs leverage the intrinsic ordering of time-series data in time. However, for non-time-series data, there does not exist a natural ordering of the dimensions. Therefore any permutation of dimensions is valid and results in a new kernel. As a result, for data with dimension $T$, $T!$ permutations exist, and performing standard cross-validation on all of those permutations quickly becomes intractable. In this paper, we only consider the \emph{default} dimension ordering and its \emph{flip} (i.e., $[1,2,3,4]$ becomes $[4,3,2,1]$), which is a common practice in the context of RNNs. We denote the configuration of a simple RNN that leverages the permutation of the input data as RNN-P. Notice that such flipping does not affect any of the kernels that are based on the BI-RNN, since by design both versions are only modeled internally. As we will see, considering only those two cases will be sufficient for RNN-P kernels to reach and surpass other methods.

\begin{figure*}[t]
    \centering 
\begin{minipage}{0.01\textwidth}
\rotatebox{90}{\;\;\;\;\;\; {\small RNN-P}}
\end{minipage}
\begin{minipage}{0.24\textwidth}
    \centering
    \subfloat{\includegraphics[width=\linewidth]{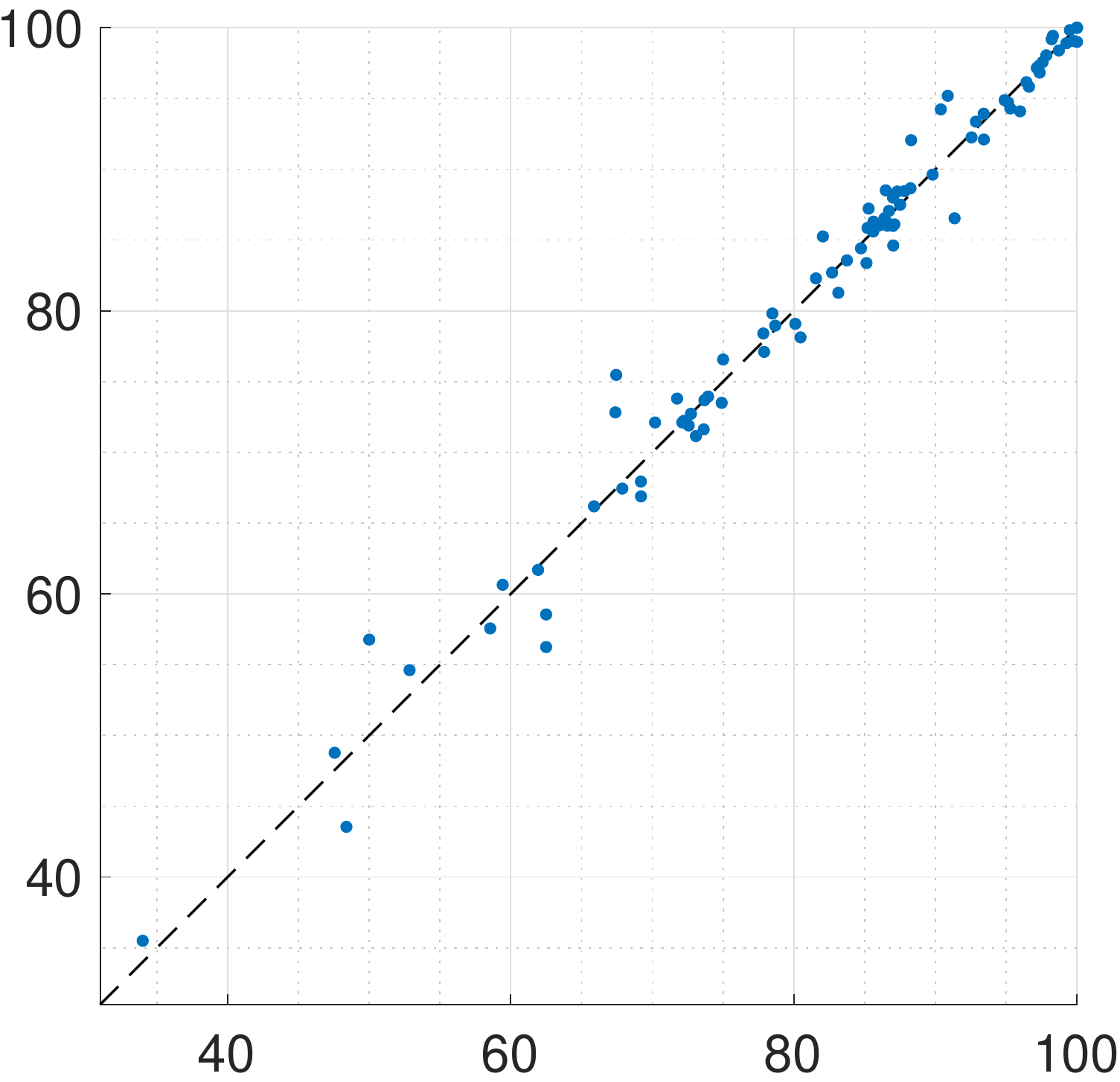} \label{fig4a} }\\
    {\vspace{1.5mm} \small\hspace{2mm} H-$\gamma$-exp} 
\end{minipage}\hfil 
\begin{minipage}{0.24\textwidth}
    \centering
    \subfloat{\includegraphics[width=\linewidth]{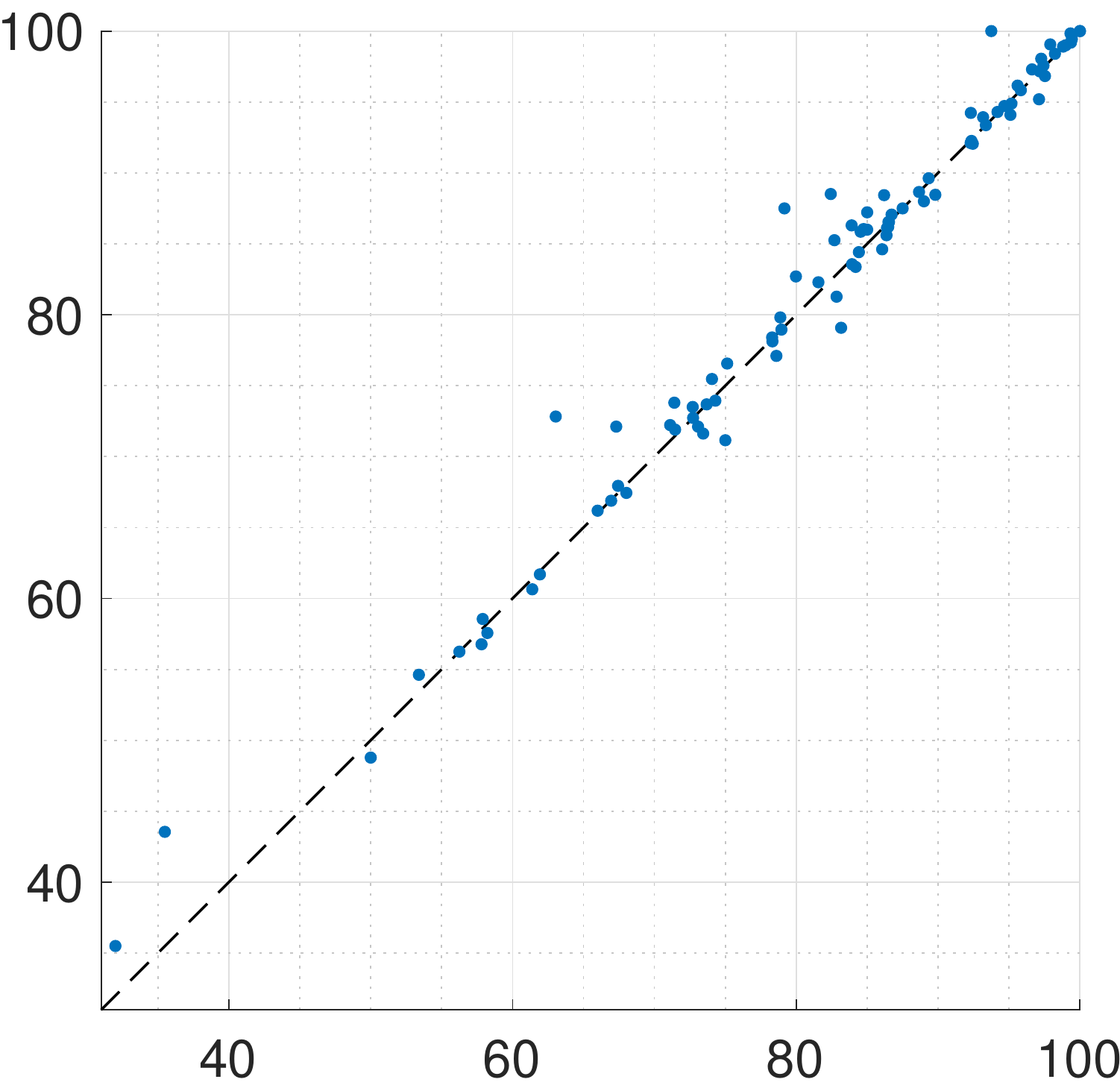} \label{fig4b}}\\
    {\vspace{1.5mm} \small \hspace{2mm} RNN-AVG} 
\end{minipage}\hfil 
\begin{minipage}{0.24\textwidth}
    \centering
    \subfloat{\includegraphics[width=\linewidth]{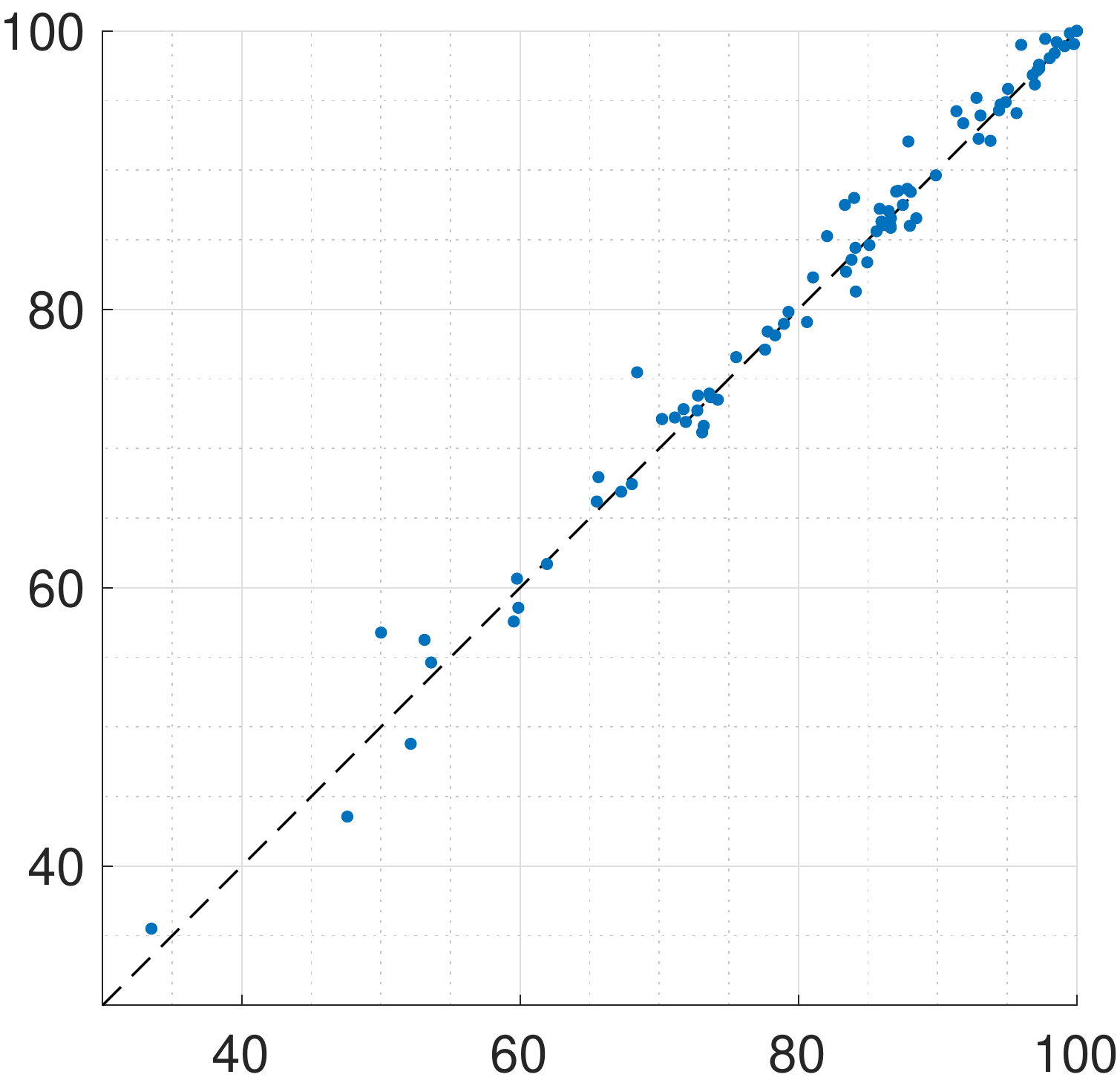} \label{fig4c}}\\ 
    {\vspace{1.5mm} \small \hspace{2mm} MLP} 
\end{minipage}
\begin{minipage}{0.24\textwidth}
    \centering
    \subfloat{\includegraphics[width=\linewidth]{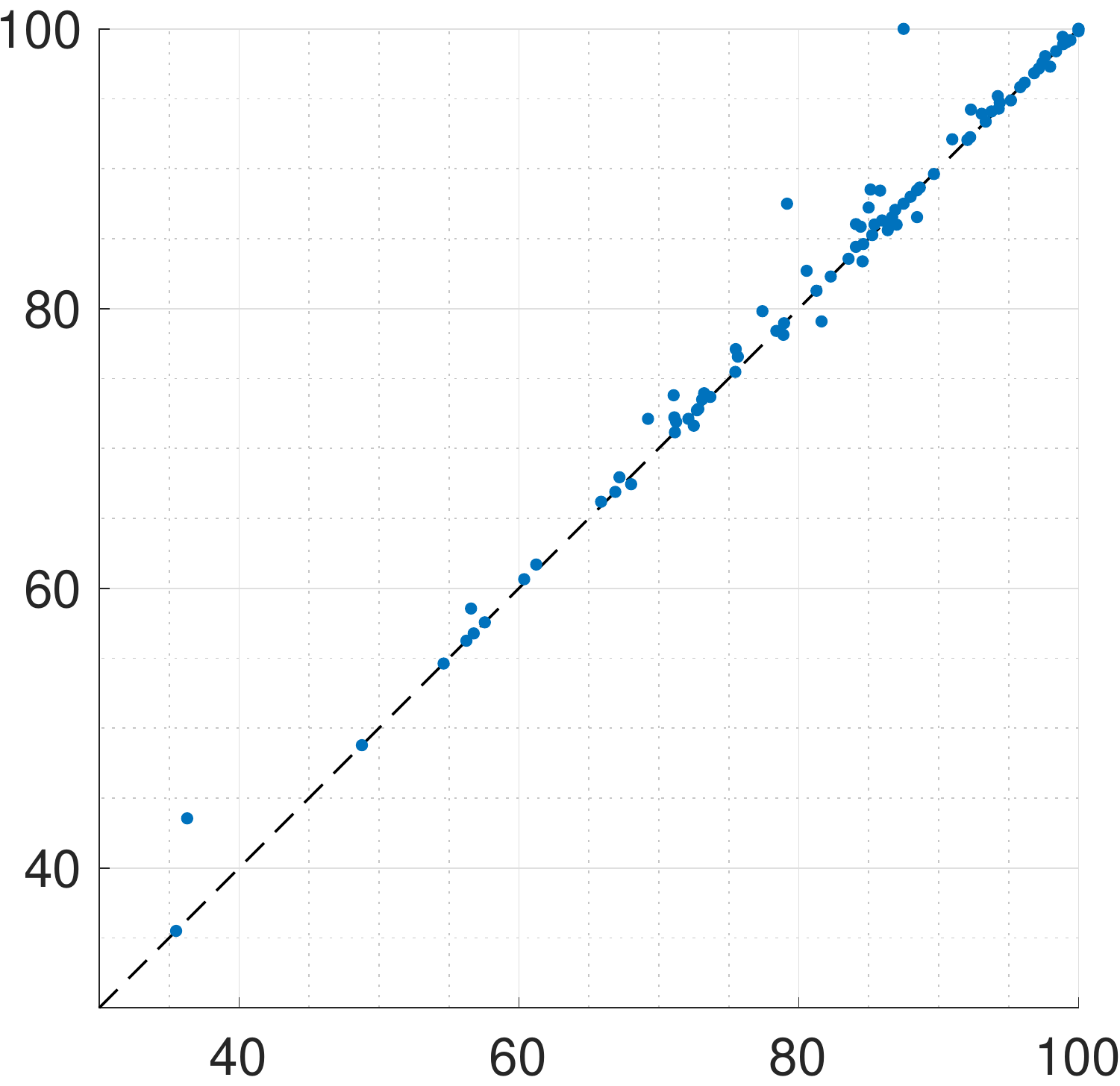} \label{fig4d}}\\
    {\vspace{1.5mm} \small \hspace{2mm} RNN} 
\end{minipage}
\caption{\small
Pairwise comparison of RNN-P with H-$\gamma$-exp, RNN-AVG, MLP, and RNN
}
\label{fig:comp}
\end{figure*}

Lastly, for each set of hyperparameters, we use \textbf{\emph{both}} CK and NTK of the different RNN architectures (simple RNN, BI-RNN, RNN-AVG, BI-RNN-AVG, RNN-P) and pick the kernel with the best performance.

It's important to consider a situation where there exists multiple hyperparameters producing the best validation performance. In this case, we incorporate all the best hyperparameters as separate models and employ voting (between those models) on the test set. This provides a generic and principled method to deal with the problem of heuristically selecting the model, in which we saw that difference between the worst and best model can result in almost half a percent average accuracy across all data sets.

We report the average test accuracy (Acc. mean) and the standard deviation (Acc. std) for all datasets. For a more detailed analysis of the results, we also calculate the PMA which is the percentage of the datasets in which a classifier achieves the maximum accuracy across all models. Additionally, we provide the P95 metric, which is the fraction of datasets in which the classifier achieves at least $95\%$ of the maximum achievable accuracy for each dataset obtained from all models being compared. Lastly, we also provide the Friedman ranking used in \cite{FERNANDEZDELGADO201911}, which is simply the average of rankings (based on Acc) of classifiers for a dataset, among all datasets. Thus, a better performing model must achieve a higher score in all the metrics except for the Friedman ranking in which case a lower score is better. 
These metrics are reported in Table \ref{tab:real-data-svm} for all the models.

\textbf{Results.} First, we observe that H-$\gamma$-exp and  BI-RNN-AVG achieve higher P95 values, demonstrating those two methods' ability to consistently produce near state-of-the-art performances. However, if we ignore the P95 metric RNN-P achieves better performance in all other metrics. The RNN-P's lower P95 can be attributed to the presence of a few datasets in which this method achieves less than $95\%$ of the best method. This might be due to the use of non-time-series data coupled with a recurrent kernel in which case the induced underlying mapping of consecutive time-steps (which is the dimension of the input) might lead to greater bias. Taking everything into account, we see significant performance gains across all other metrics for RNN-P, which should motivate practitioners to consider RNN-P as a novel baseline kernel even for non-time-series data.

To visualize the comparison between RNN-P and the top 4 classifiers in terms of the average test accuracy, we present pairwise scatter plot of the test accuracy for all datasets in Figure \ref{fig:comp}.

\section{Conclusion And Discussion}

In this paper, we extended the NTK for RNNs to other variants of recurrent architectures such as bi-directional RNNs and (bi-directional) RNNs with average pooling. We have provided the code to compute those kernels on CPU and GPU and tested the performance of the kernels induced by infinite width recurrent architectures on 90 datasets of non-time-series data from the UCI data repository, where our proposed kernels outperforms alternative methods.
We believe that the reported empirical results on non-time-series datasets along with the provided accessible and fast implementation should push practitioners to use kernels induced by recurrent architecture on datasets of any kind. In fact, now incorporating those derived kernels as part of the rich set of existing NTKs should provide enough diversity in available kernels to truly provide performances going beyond traditional methods such as random forest on any given dataset.

An interesting future direction would be to combine those newly obtained kernels in applications such as similarity measure of data with different dimensionality. This has already been explored when employing the CK and NTK of simple RNNs \cite{alemohammad2020recurrent, yang2020tensor, yang2019tensor,alemohammad2020wearing}.

\section*{Acknowledgments}
SA, ZW and RichB were supported by NSF grants CCF-1911094, IIS-1838177, and IIS-1730574; ONR grants N00014-18-12571, N00014-20-1-2534, and MURI N00014-20-1-2787; AFOSR grant FA9550-18-1-0478; and a Vannevar Bush Faculty Fellowship, ONR grant N00014-18-1-2047.
\small
\bibliography{ref}
\bibliographystyle{ieeetr}

\end{document}